\documentclass{article}

\usepackage{amsmath}  
\usepackage{enumitem}  
\usepackage{pdfpages} 

\usepackage{times}
\usepackage{graphicx} 
\usepackage{subfigure} 

\usepackage{natbib}

\usepackage{algorithm}
\usepackage{algorithmic}

\usepackage{hyperref}

\usepackage[accepted]{icml2018}

\icmltitlerunning{Not to Cry Wolf: Distantly Supervised Multitask Learning in Critical Care}

\begin{document} 

\twocolumn[
\icmltitle{Not to Cry Wolf:\\ Distantly Supervised Multitask Learning in Critical Care}




\begin{icmlauthorlist}
\icmlauthor{Patrick Schwab}{eth1}
\icmlauthor{Emanuela Keller}{usz}
\icmlauthor{Carl Muroi}{usz}
\icmlauthor{David J. Mack}{usz}
\icmlauthor{Christian Str\"assle}{usz}
\icmlauthor{Walter Karlen}{eth1}
\end{icmlauthorlist}

\icmlaffiliation{eth1}{Institute of Robotics and Intelligent Systems, ETH Zurich, Switzerland}
\icmlaffiliation{usz}{Neurocritical Care Unit, Department of Neurosurgery, University Hospital Zurich, Switzerland}

\icmlcorrespondingauthor{Patrick Schwab}{patrick.schwab@hest.ethz.ch}

\icmlkeywords{boring formatting information, machine learning, ICML}

\vskip 0.3in
]



\printAffiliationsAndNotice{}  

\newcommand{\themethod}[0]{DSMT-Net}

\begin{abstract}
Patients in the intensive care unit (ICU) require constant and close supervision. To assist clinical staff in this task, hospitals use monitoring systems that trigger audiovisual alarms if their algorithms indicate that a patient's condition may be worsening. However, current monitoring systems are extremely sensitive to movement artefacts and technical errors. As a result, they typically trigger hundreds to thousands of false alarms per patient per day - drowning the important alarms in noise and adding to the exhaustion of clinical staff. In this setting, data is abundantly available, but obtaining trustworthy annotations by experts is laborious and expensive. We frame the problem of false alarm reduction from multivariate time series as a machine-learning task and address it with a novel multitask network architecture that utilises distant supervision through multiple related auxiliary tasks in order to reduce the number of expensive labels required for training. We show that our approach leads to significant improvements over several state-of-the-art baselines on real-world ICU data and provide new insights on the importance of task selection and architectural choices in distantly supervised multitask learning.
\end{abstract}
\vskip -0.35in
\begin{figure}[ht!]
\begin{center}
\centerline{\includegraphics[scale=0.50]{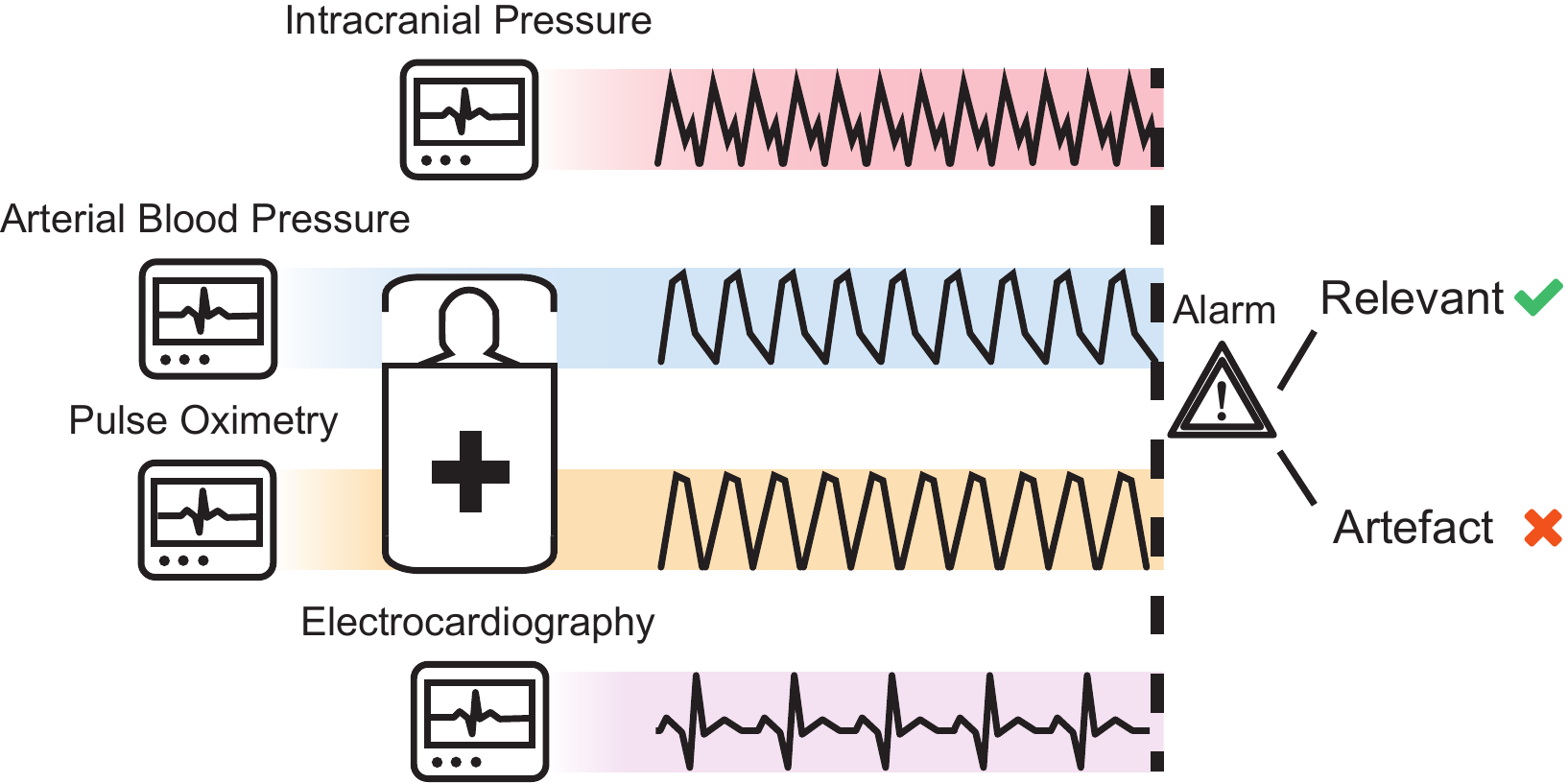}}
\caption{Schematic overview of the described problem setting in critical care. Before an alarm is brought to the attention of clinical staff, an alarm classification algorithm could analyse a recent window of the full set of available data streams in order to identify whether the alarm was likely caused by an artefact or technical error and may therefore be reported with a lower degree of urgency.}
\label{fig:overview}
\end{center}
\vskip -0.65in
\end{figure}
\section{Introduction}
False alarms are an enormous mental burden for clinical staff and are extremely dangerous to patients, as alarm fatigue and desensitisation may lead to clinically important alarms being missed \cite{drew2014insights}. Reportedly, several hundreds of deaths a year are associated with false alarms in patient monitoring in the United States alone \cite{cvach2012monitor}. \\

An intelligent alarm classification system could potentially reduce the burden of a large subset of those false alarms by assessing which alarms were likely caused by either an artefact or a technical error and reporting those alarms with a lower degree of urgency (Figure \ref{fig:overview}). Roadblocks that have so far prevented the adoption of machine learning for this task are the heterogeneity of monitoring systems, the requirement for an extremely high specificity, to avoid suppressing important alarms, and the prohibitively high cost associated with obtaining a representative set of clinically validated labels for each of the manifold alarm types and monitoring system configurations in use at hospitals.

We present a semi-supervised approach to false alarm reduction that automatically identifies and incorporates a large amount of distantly supervised auxiliary tasks in order to significantly reduce the number of expensive labels required for training. We demonstrate, on real-world ICU data, that our approach is able to correctly classify alarms originating from artefacts and technical errors better than several state-of-the-art methods for semi-supervised learning when using just 25, 50 and 100 labelled samples. Besides their importance for clinical practice, our results highlight the power of distant multitask supervision as a flexible and effective tool for learning when unlabelled data are readily available, and shed new light on semi-supervised learning beyond low-resolution image benchmark datasets. 

\textbf{Contributions.} We subdivide this work along the following distinct contributions:
\begin{itemize}[noitemsep]
\item[(i)] We introduce \themethod s: A novel neural architecture built on the idea of utilising distant supervision through multiple auxiliary tasks in order to better harness unlabelled data.
\item[(ii)] We present a methodology for selecting a large set of related auxiliary tasks in time series, and a training procedure that counteracts adverse gradient interactions between auxiliary tasks and the main task.
\item[(iii)] We perform extensive quantitative experiments on a real-world ICU dataset consisting of almost 14,000 alarms in order to evaluate the relative classification performance and label efficiency of \themethod s compared to several state-of-the-art methods. 
\end{itemize}
\section{Related Work}
\textbf{Background.} Driven by widespread efforts to automate patient monitoring, there has been a recent surge in works applying machine learning to the vast amounts of data generated in ICUs. One notable driver is the MIMIC \cite{saeed2011multiparameter} dataset that has made ICU data accessible to a large number of researchers. Related works have, for example, explored the use of ICU data for tasks such as mortality modelling \cite{ghassemi2014unfolding}, illness assessment and forecasting \cite{ghassemi2015multivariate}, diagnostic support \cite{lipton2015learning}, patient state prediction \cite{cheng2017sparse} and learning weaning policies for mechanical ventilation \cite{prasad2017reinforcement}. Applying machine-learning approaches to clinical and physiological data is challenging, because it is heterogenous, noisy, confounded, sparse and of high temporal resolution over long periods of time. These properties are in stark contrast to many of the benchmark datasets that machine-learning approaches are typically developed and evaluated on. Several works therefore deal with adapting existing machine-learning approaches to the idiosyncrasies of clinical and physiological data, such as missingness \cite{lipton2016directly,che2016recurrent}, long-term temporal dependencies \cite{choi2016doctor}, noise \cite{schwab2017beat}, heterogeneity \cite{libbrecht2015machine} and sparsity \cite{lasko2013computational}. We build on several of these innovations in this work.

\textbf{Alarm Fatigue.} The PhysioNet 2015 challenge on false alarm reduction in electrocardiography (ECG) monitoring \cite{clifford2015physionet} was one of the most notable efforts to date to address the issue of false alarms in physiological monitoring. Within the challenge, researchers introduced several effective approaches to reducing the false alarm rate of arrhythmia alerts in ECGs \cite{fallet2015multimodal,eerikainen2015decreasing,krasteva2016real,plesinger2016taming}. However, clinicians in the ICU do not just monitor for arrhythmias, but many adverse events at once using a multitude of different monitoring systems. Typically, these monitoring systems operate in isolation on a single biosignal and trigger their own distinct sets of alarms. Previous research has shown that there is an opportunity to use data from other biosignals to identify false alarms in related waveforms \cite{aboukhalil2008reducing}. We therefore believe that a comprehensive solution to alarm fatigue requires an approach that accounts for the monitoring setup as a whole, rather than targeting specific systems or alarms in isolation. 

\textbf{Distant Supervision and Multitask Learning.} Multitask learning has a rich history in healthcare applications and has, for example, been used for risk prediction in neonatal intensive care \cite{saria2010integration}, drug discovery \cite{ramsundar2015massively} and prediction of Clostridium difficile \cite{wiens2016patient}. A way of leveraging multitask learning to improve label-efficiency is to learn jointly from complementary unsupervised auxiliary tasks along with the supervised main task. Existing literature refers to the concept of applying indirect supervision through auxiliary tasks, be it for label-efficiency or additional predictive performance, as weak supervision \cite{papandreou2015weakly,oquab2015object}, distant supervision \cite{zeng2015distant,deriu2017leveraging} or self-supervision \cite{fernando2017self,doersch2017multi}. In particular, \cite{doersch2017multi} used distantly supervised multitask learning to increase predictive performance in computer vision with up to four hand-engineered auxiliary tasks. Using auxiliary tasks in addition to a main task has also been shown to be a promising approach in reinforcement \cite{jaderberg2016reinforcement,aytar2018playing} and adversarial learning \cite{salimans2016improved}. Recently, \cite{laine2016temporal} proposed to use outputs from the same model at different points in training and with varying amounts of regularisation as additional unsupervised targets for the main task. In contrast to existing works, we present the first approach to distantly supervised multitask learning that automatically identifies a large set of related auxiliary tasks from multivariate time series to jointly learn from labelled and unlabelled data. In addition, our approach scales to hundreds of auxiliary tasks in an end-to-end trained neural network.

\textbf{Semi-supervised Learning.} Beside distant supervision, other state-of-the-art approaches to semi-supervised learning in neural networks include, broadly, methods based on (i) reconstruction objectives, such as Variational Auto-Encoders (VAEs) \cite{kingma2014semi} and Ladder Networks \cite{rasmus2015semi}, and (ii) adversarial learning \cite{springenberg2015unsupervised,dai2017good,li2017triple}. However, with standard benchmarks consisting primarily of low-resolution image datasets, it is yet unclear to what degree these method's results generalise to heterogenous, long-term and high-resolution time series datasets with informative missingness, as commonly encountered in healthcare applications.
\begin{figure*}[ht!]
\vskip -0.05in
\begin{center}
\subfigure[\themethod-0]{\includegraphics[scale=0.75]{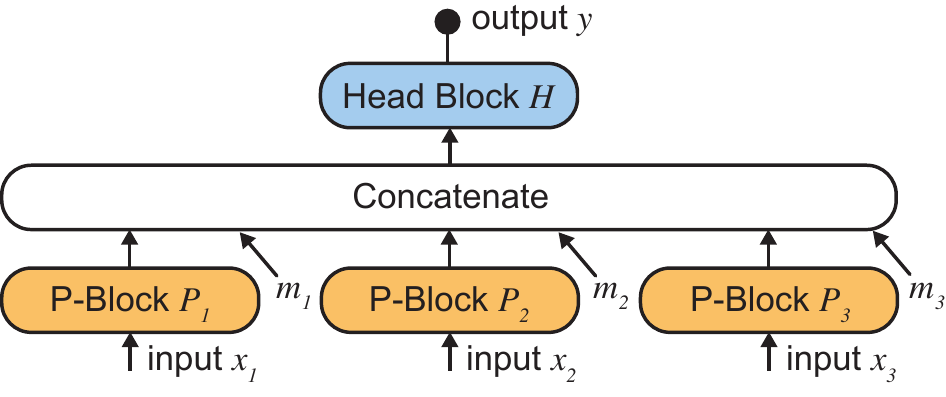}}\quad
\subfigure[\themethod-3]{\includegraphics[scale=0.75]{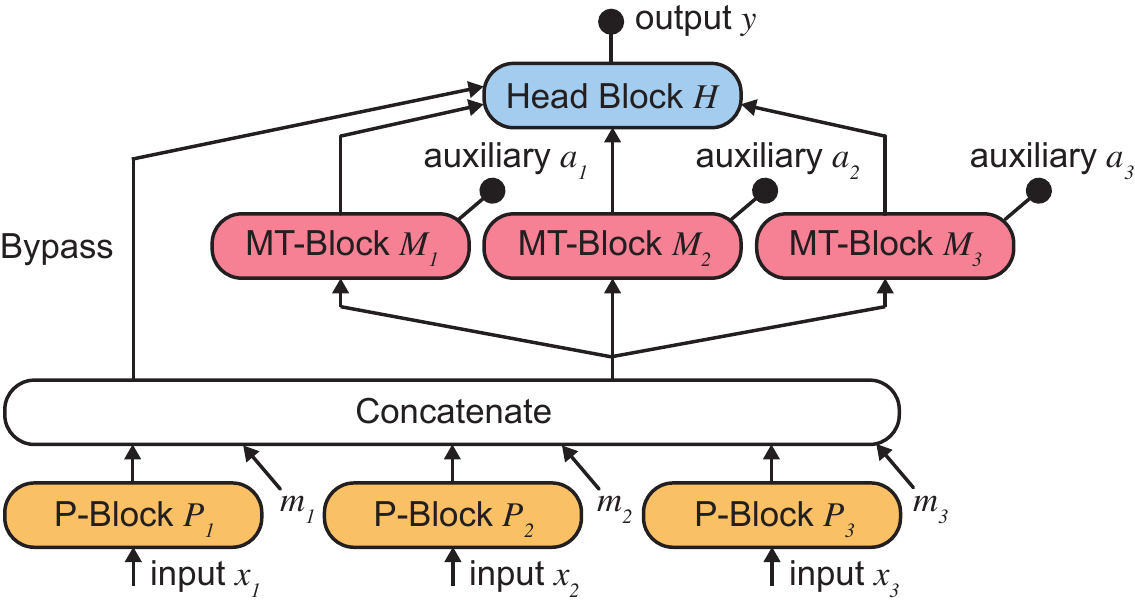}}\quad
\caption{Two \themethod{} architectures: (a) one without (\themethod-0) and (b) one with three (\themethod-3) auxiliary tasks. The number of horizontally aligned multitask blocks $M_j$ (MT-Blocks; red) is variable. Each multitask block hosts its own auxiliary task $a_j$. An additional bypass connection gives the head block $H$ (blue) direct access to the concatenated hidden states of the perception blocks $P_i$ (P-Blocks; orange). Each perception block operates on its own input data stream $x_i$. The model incorporates binary missing indicators $m_i$ for each perception block to handle situations where input data streams are missing. }
\label{fig:architecture}
\end{center}
\vskip -0.2in
\end{figure*}

\section{Distantly Supervised Multitask Networks}
\label{sec:dsmt}
Distantly Supervised Multitask Networks (\themethod s) are end-to-end trained neural networks that process $n$ heterogenous input data streams $x_i$ in order to solve a multitask learning problem with one main task and $k$ auxiliary tasks designed to augment the main task. Conceptually, a \themethod{} consists of the following components: One perception block $P_i$ with $i\in[1,n]$ for each of the $n$ input data streams $x_i$, a variable number $k$ of multitask blocks $M_j$ with $j\in[1,k]$, and a single head block $H$ (Figure \ref{fig:architecture}b). Each of these block types is itself a neural network with its own parameters and arbitrary architectures and hyperparameters. The role of the perception blocks $P_i$ is to extract a hidden feature representation $h_{p,i}$ from their respective input data streams $x_i$:
\vskip -0.2in
\begin{align}
\label{eq:hpi} h_{p,i} &= P_i(x_i)
\end{align}
\vskip -0.05in
We separate the perception blocks by input stream $x_i$ in order to be able to model a dynamic set of potentially missing input data streams. To allow our model to learn missingness patterns, we follow \cite{lipton2016directly} and accompany each perception block with a missing indicator $m_i$ that is set to $0$ if the data stream $x_i$ is present and $1$ if it is missing. We additionally perform zero-imputation on the missing perception blocks' features $h_{p,i}$.
We then concatenate the features $h_{p,i}$ extracted from the perception blocks and the corresponding missing indicators $m_i$ into a joint feature representation $P_c$ over all input data streams: 
\vskip -0.2in
\begin{align}
\label{eq:pc} P_c &= \text{concatenate}(h_{p,1}, m_1, ..., h_{p,n}, m_n)
\end{align}
\vskip -0.05in
The joint feature representation $P_c$ combines the information from all feature representations of the input data streams and serves as input to the higher level multitask blocks and the head block. The main role of multitask blocks $M_j$ is to host auxiliary tasks $a_j$. All multitask blocks are aligned in parallel in order to minimise the distance gradients have to propagate through both to the joint feature representation $P_c$ and from the head block. As output, each multitask block produces a hidden high-level feature representation $h_{m,j}$:

\vskip -0.2in
\begin{align}
\label{eq:hmj}
h_{m,j} &= M_j(P_c) 
\end{align}
\vskip -0.05in

Compared to the straightforward approach of directly appending the auxiliary tasks to the head block $H$, the positioning of multitask blocks below the head block achieves separation of concerns. In \themethod s, the head block focuses on learning a hidden feature representation that is optimised solely for the main task rather than being forced to learn a joint feature representation that performs well on multiple, possibly competing tasks. 

The head block $H$ computes the final model output $y$ and further processes the hidden feature representations $h_{m,j}$ of the multitask blocks via a combinator function (equation (\ref{eq:y})). In addition to the hidden feature representations of the multitask blocks, the head block retains direct access to $P_c$ via a bypass connection. We motivate the inclusion of a bypass connection with the desire to learn hidden feature representations in multitask blocks that add information over $P_c$ \cite{he2016deep}. Mathematically, we formulate the head block $H$ as follows:

\vskip -0.2in
\begin{align}
\label{eq:y}
y &= H(\text{combine}_{\text{MLP}}(P_c, h_{m,1}, ..., h_{m,k}))
\end{align}
\vskip -0.05in

We note that the \themethod{} architecture without any multitask blocks corresponds to a na\"{i}ve supervised neural network over a mixture of expert networks \cite{jordan1994hierarchical,shazeer2017outrageously,schwab2018granger} for each input data stream $x_i$ (\themethod-0; Figure \ref{fig:architecture}a).

\textbf{Combinator Function.} In \themethod s, the combinator function integrates $m+1$ data flows from the $m$ multitask blocks' hidden representations as well as the joint feature representation $P_c$. We propose a combinator function ($\text{combine}_\text{MLP}$) that consists of a single hidden-layer multi-layer perceptron (MLP) with a dimensionality twice as big as a single multitask block's feature representation. As input, the MLP receives the concatenation of all the feature representations to be integrated:
\vskip -0.2in
\begin{align}
\label{eq:y}
\text{combine}_\text{MLP} &= \text{MLP}(\text{concatenate}(P_c, h_1, ..., h_m))
\end{align}
\vskip -0.05in

\subsection{Selection of Auxiliary Tasks}

One of the most important questions in distantly supervised learning is how to identify suitable auxiliary tasks. A common choice of auxiliary task for un- and semi-supervised learning is reconstruction over the feature and/or hidden representation space. Several modern semi-supervised methods take this approach \cite{vincent2008extracting,kingma2014auto,kingma2014semi,rasmus2015semi}. Reconstruction is a convenient choice of auxiliary task because it is generically applicable to any input data, neural architecture and predictive task. However, given recent empirical successes by distant supervision with specifically engineered auxiliary tasks \cite{oquab2015object,deriu2017leveraging,doersch2017multi}, we reason that (i) more "related" tasks might be a better choice of auxiliary task for semi-supervised learning than reconstruction \cite{ben2003exploiting} and that (ii) using multiple diverse auxiliary tasks might be more effective than just one \cite{baxter2000model}. Since a predictive feature for a main task is also a good auxiliary task for learning shared predictive representations \cite{ando2005framework}, we follow a simple two-step feature selection methodology \cite{christ2016distributed} to automatically identify a large set of auxiliary tasks that are closely related to the main task:
\begin{itemize}[noitemsep]
\item[1.] We extract features from a large pool of manually-designed features from each input time series. Due to the large wealth of research in manual feature engineering, there exist vast repositories of such features for many data modalities, e.g. \cite{christ2016distributed}. For time series, examples of such features would be, e.g., the autocorrelation at different lag levels or the power spectral density over a specific frequency range.
\item[2.] We statistically test the extracted features for their importance related to the main task in order to rank the features by their estimated predictive potential and determine their relevance. A suitable statistical test is, for example, a hypothesis test for correlation between the labels $y_{true}$ and the extracted features using Kendall's $\tau$ \cite{kendall1945treatment}.
\end{itemize}
\vskip -0.15in
Using this approach, we are able to identify a large, ranked list of predictive features suitable for use as target labels for auxiliary tasks $a_j$ in \themethod s. There are two approaches to choosing a subset of those features as auxiliary targets: (i) in order of feature importance or (ii) randomly out of the set of relevant features. The main difference between the two approaches is that random selection has a higher expected task diversity as similar tasks are likely to also rank similarly in terms of importance. There are arguments both for (more information per task) and against (harder to learn shared feature representation) higher task diversity. We therefore evaluate both approaches in our experiments.

\subsection{Training Distantly Supervised Multitask Networks}
\label{sec:improved_training}

A key problem when training neural networks on multiple tasks simultaneously using stochastic gradient descent is that gradients from the different tasks can interfere adversely \cite{teh2017distral,doersch2017multi}. We therefore completely disentangle the training of the unsupervised and supervised tasks in \themethod s. Instead of training the auxiliary tasks jointly with the main task, we alternate between optimising \themethod s for the auxiliary tasks and the main task in each epoch, starting with the auxiliary tasks. At the computational cost of an additional pass during training, the two-step training procedure prevents any potential adverse intra-step gradient interactions between the two classes of tasks. To ensure similar convergence rates for both the main and auxiliary tasks, we weight the auxiliary tasks such that the total learning rate for the unsupervised and supervised step are approximately the same, i.e. a weight of $\frac{1}{k}$ for each auxiliary task when there are $k$ auxiliary tasks. A similar training schedule, where generator and discriminator networks are trained one after another in each iteration, has been proposed to train generative adversarial networks (GANs) \cite{goodfellow2014generative}. 

\section{Experiments}
We performed extensive quantitative experiments\footnote{The source code for this work is available online at \\\url{https://github.com/d909b/DSMT-Nets}.} on real-world ICU data using a multitude of different hyperparameter settings in order to answer the following  questions:
\begin{itemize}[noitemsep]
\item[(1)] \textit{How do \themethod s perform in terms of predictive performance and label efficiency in multivariate false alarm detection relative to state-of-the-art methods for semi-supervised learning?}
\item[(2)] \textit{What is the relationship between the number of auxiliary tasks, predictive performance and label efficiency?}
\item[(3)] \textit{What is the importance of the architectural separation of auxiliary tasks and the main task and the two-step training procedure in \themethod s?}
\item[(4)] \textit{Is there value in selecting a specific set of related auxiliary tasks for distantly supervised multitask learning over random selection?}
\end{itemize}
\vskip -0.15in
To answer question (1), we systematically evaluated \themethod s and several baseline models in terms of their area under the receiver operator curve (AUROC) using varying amounts of manually classified labels $n_{\text{labels}} = (12, 25, 50, 100, 500, 1244)$ and varying amounts of auxiliary tasks $k = (6, 12, 25, 50, 100)$ in the \themethod s. We chose the label subsets at random without stratification. The comparison between the models' performances when using different levels of labels enable us to judge the label efficiency of the compared models, i.e. which level of predictive performance they can achieve with a limited amount of labels. By also changing the amount of auxiliary tasks used in the models, we are additionally able to assess the relationship of the number of auxiliary tasks with label efficiency and predictive performance (question (2)).

To answer question (3), we performed an ablation study using the \themethod s with 100 auxiliary tasks (\themethod-100) using varying amounts of manually classified labels as base models. We then trained the same models without the two-step training procedure (- two step train). In addition, we evaluated the performance of a deep Highway Network \cite{srivastava2015training} with the same 100 auxiliary tasks distributed sequentially among layers (\themethod-100D) to compare multitask learning in depth against width. Lastly, we also evaluated a multitask network where the same 100 auxiliary tasks are placed directly on the head block (Na\"{i}ve Multitask Network). Through this process, we aimed to determine the relative importance of the individual design choices introduced in section \ref{sec:dsmt}.

To answer question (4), we compared the predictive performance of \themethod s using a random selection of all the significant features as determined by our feature selection methodology to that of \themethod s that use a selection in order of feature importance. We do so with \themethod s with 6 (\themethod-6R) and 100 (\themethod-100R) auxiliary tasks to additionally assess whether the importance of auxiliary task selection is sensitive to the number of auxiliary tasks. 

In total, we trained 2730 distinct model configurations in order to gain a better understanding of the empirical strengths and weaknesses of \themethod s.

\begin{figure}[ht!]
\begin{center}
\centerline{\includegraphics[width=\columnwidth]{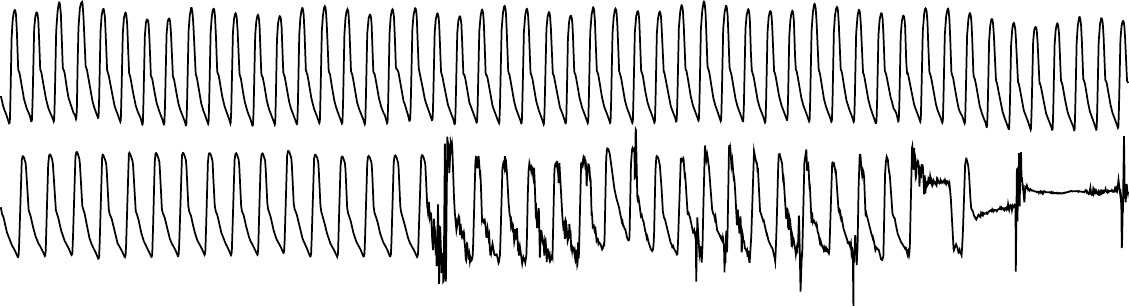}}
\vskip -0.15in
\caption{Two clear examples of arterial blood pressure signals without (top) and with pronounced artefacts (bottom). Note the high frequency noise and atypical  shape in the artefact sample.}
\label{fig:artefact}
\end{center}
\vskip -0.45in
\end{figure}

\subsection{Dataset}
We collected biosignal monitoring data from January to August 2017 (8 months) from consenting patients admitted to the Neurocritical Care Unit at the University of Zurich, Switzerland. The data included continuous, evenly-sampled waveforms obtained by electrocardiography (ECG; 200 Hz), arterial blood pressure (ART; 100 Hz), pulse oximetry (PPG and SpO$_2$; 100 Hz) and intracranial pressure (ICP; 100 Hz) measurements. For this study, we did not collect or make use of any personal, demographic or clinical data, such as prior diagnoses, treatments or electronic health records. To obtain a ground truth assessment of alarms, we provided clinical staff with a user interface and instructions\footnote{Detailed instructions and full qualitative samples can be found in the supplementary material.} for annotating alarms that they believed were caused by artefacts or a technical error (Figure \ref{fig:artefact}). Because of technical limitations in exporting data from the biosignal database, we selected the subset of 20 monitoring days of 14 patients with the highest amount of manually labelled alarms for further analysis. The evaluated dataset encompassed a grand total of 13,938 alarms, yielding an average rate of $696.9$ alarms per patient per day. This number is in line with alarm rates reported in previous works \cite{cvach2012monitor}. Of all alarms, $46.99\%$ were caused by an alarm-generating algorithm operating on the ART waveform, $33.10\%$ on PPG or derived SpO$_2$, $12.02\%$ on ICP and $7.89\%$ on either ECG or an ECG-derived signal, such as the heart rate signal. 

\textbf{Annotations.} Out of the whole set of alarms, 1,777 ($12.75\%$) alarms were manually labelled by clinical staff during the observed period. Because we used multiple annotators that were not calibrated to each other's assessments, we additionally conducted a review over all 1,777 annotations in order to ensure the internal consistency of the set of annotations as a whole. In this review round, we found that a total of 603 (33.93\%) annotations were inconsistent. We subsequently assigned corrected labels to these alarms. Since label quality is paramount for model training and validation, we suggest at least one label review round using the majority vote of a committee of labellers with clear instructions in order to maintain a sufficient degree of label consistency. Recent large-scale labelling efforts in physiological monitoring of arrhythmias \cite{clifford2017af} suggest that even more review rounds might be necessary to obtain a gold standard set of labels. In our final label set, 976 (45.08\%) out of all annotated alarms are labelled as most likely being caused by an artefact or technical error. We note that a data collection effort of this scale is extremely expensive and therefore economically infeasible for most hospitals, motivating our search for a more label-efficient approach.

\subsection{Evaluation Setup}
As input data, we extracted a $40$ second window of the time frame immediately before an alarm was triggered from each available biosignal. We considered a signal stream to be missing for a given alarm setting if the last recorded measurement of that type happened longer than 10 seconds ago. To reduce the computational resources required for our experiments, we resampled the input data to $\frac{1}{16}^{th}$ of its original sampling rate. In our preliminary evaluation, we did not see significant performance changes when using a higher sampling rate or a longer context window. Additionally, we standardised the extracted windows of each stream to the range of $[-1,1]$ using the maximum and minimum values encountered in that window.

\textbf{Baselines.} To ensure a fair reference, we used the \themethod s base architecture without any horizontal blocks and auxiliary tasks as the supervised baseline (\themethod-0, Figure \ref{fig:architecture}a). Because the supervised baselines have no auxiliary tasks, we trained them in a purely supervised manner on the labelled alarms only.

As a baseline for feature selection, we used the automated feature extraction and selection approach from \cite{christ2016distributed} to identify a large number (up to 875) of relevant time series features from the multivariate input data. Note that we followed this process separately for each distinct amount of labels in order to avoid information leakage. We then fed those features to a random forest (RF) classifier consisting of $4096$ trees to produce predictions (Feature RF). As mentioned in section \ref{sec:improved_training}, we used the same feature selection approach to identify suitable auxiliary tasks for \themethod s. The Feature RF baseline therefore serves as a reference for directly using the identified significant features to make a prediction.

For comparison to the state-of-the-art in reconstruction-based semi-supervised learning, we evaluated Ladder Networks \cite{rasmus2015semi} on the same dataset. We replaced the \themethod{} components on top of the joint feature representation $P_c$ with a Ladder Network in order to use a comparable architecture that is also able to model missingness and heterogenous input streams.

For comparison to the state-of-the-art in semi-supervised adversarial learning, we trained GANs using a semi-supervised objective function and feature matching \cite{salimans2016improved} on the same dataset. This type of GAN has been shown to be highly efficacious at semi-supervised learning in low-resolution image datasets \cite{salimans2016improved}. We trained the generator networks to generate a context window of multiple high-resolution time series as input to a \themethod{} discriminator without any auxiliary tasks. In terms of architecture, the generator networks used strided upsampling convolutions.

\textbf{Hyperparameters.} To ensure a fair comparison, we used a systematic approach to hyperparameter selection for each evaluated neural network. We trained each model 35 times with a random choice of the three variable hyperparameters bound to the same ranges ($1-3$ hidden layers, $16-32$ units/filters per hidden layer, $25\% - 85\%$ dropout). We reset the random seed to the same value for each model in order to make the search deterministic across training runs, i.e. all the models were evaluated on exactly the same set of hyperparameter values. Note that this setup does not guarantee optimality for any model, however, with respect to the evaluated hyperparameters, it guarantees the models were evaluated fairly and given the same amount of scrutiny. To train the neural network models, we used a learning rate of $0.001$ for the first ten epochs and $0.0001$ afterwards to optimise the binary cross-entropy for the main classification output and the mean squared error for all auxiliary tasks. We additionally used early stopping with a patience of 13 epochs. For the extra hyperparameters in Ladder Networks, we set the noise level to be fixed at $0.2$ at every layer, the denoising loss weight to $100$ for the first hidden layer and to $0.1$ for every following hidden layer. For the GAN models, we used a base learning rate of $0.0003$ for the discriminator and a slightly increased learning rate of $0.003$ for the generator to counteract the faster convergence of the discriminator networks. We trained GANs using an early stopping patience on the main loss of 650 steps for a minimum of 2500 steps. To choose these extra hyperparameters of GANs and Ladder Networks, we followed the original author's published configurations \cite{rasmus2015semi,salimans2016improved} and adjusted them slightly to ensure they converged.

\begin{table*}[ht!]
\vskip -0.1in
\caption{Comparison of the maximum AUROC value across the 35 distinct models (vertical) that we trained using different sets of hyperparameters and varying amounts of labels (horizontal). We report the AUROC of the best encountered model as calculated on the test set of 533 alarms. The best results in each column are highlighted in bold.}
\vskip 0.1in
\label{tb:results}
\centering
\begin{small}
\hskip -0.03in
\begin{tabular}{l*{6}{r}}
AUROC with \# of Labels & 12 & 25 & 50 & 100 & 500 & 1244 \\
\hline
Feature RF & \hspace{1cm} 0.567 & \hspace{0.5cm} 0.574 & \hspace{0.5cm} 0.628 & \hspace{0.5cm} 0.822 & \hspace{0.5cm} \textbf{0.942} & \hspace{0.5cm} \textbf{0.955} \\ 
Supervised baseline & 0.751 & 0.753 & 0.806 & 0.873 & 0.941 & 0.942\\
Na\"{i}ve Multitask Network & 0.791 & 0.804 & 0.828 & 0.887 & 0.941 & 0.940\\
Ladder Network & 0.791 & 0.772 & 0.800 & 0.842 & 0.863 & 0.868\\
Feature Matching GAN & \textbf{0.846} & 0.834 & 0.834 & 0.865 & 0.911 & 0.898\\ \\
\themethod-6 & 0.763 & 0.839 & 0.866 & 0.897 & 0.924 & 0.934\\
\themethod-12 & 0.739 & \textbf{0.872} & 0.891 & 0.890 & 0.928 & 0.933\\
\themethod-25 & 0.761 & 0.870 & 0.886 & 0.898 & 0.924 & 0.929\\
\themethod-50 & 0.722 & 0.847 & \textbf{0.901} & 0.906 & 0.926 & 0.936\\
\themethod-100 & 0.720 & 0.831 & 0.893 & 0.907 & 0.934 & 0.934\\
- two step train & 0.733 & 0.798 & 0.785 & 0.814 & 0.849 & 0.898\\ \\
\themethod-6R & 0.805 & 0.851 & 0.884 & \textbf{0.909} & 0.921 & 0.938\\
\themethod-100R & 0.790 & 0.860 & 0.883 & \textbf{0.909} & 0.918 & 0.932\\
\themethod-100D & 0.587 & 0.611 & 0.722 & 0.610 & 0.624 & 0.702\\
\end{tabular}
\end{small}
\end{table*}
\textbf{Architectures.} We used the conceptual architecture from Figure \ref{fig:architecture} as a base architecture for the \themethod s. As perception blocks, we employed ResNets \cite{he2016deep} with 1-dimensional convolutions over the time axis for each input data stream. As head block and multitask blocks, we used Highway Networks \cite{srivastava2015training}. The head block hosted a sigmoid binary output $y$ that indicated whether or not the proposed alarm was likely caused by an artefact. In addition, we used batch normalisation in the \themethod{} blocks.

\textbf{Metrics.} For each approach, we report the AUROC of the best model encountered over all 35 hyperparameter runs.

\textbf{Dataset Split.} We applied a random split stratified by alarm classification to the whole set of annotated alarms to separate the available data into a training ($70\%$, 1244 alarms) and test set ($30\%$, 533 alarms). 

\section{Results and Discussion}
We report the results of our experiments in Table \ref{tb:results} and discuss them in the following paragraphs.

\textbf{Predictive Performance.} Overall, we found that the label limit after which the purely supervised approaches consistently outperformed the semi-supervised approaches was between 100 and 500 labels. The strongest approach when using all 1244 available labels and the 500 label subset was the purely supervised Feature RF baseline. Out of all compared methods, \themethod s were the most label-efficient approach when using 25, 50 and 100 labels. However, the Feature Matching GAN outperformed the \themethod s when using just 12 labels.  In our experimental setting, the best \themethod s yielded significant improvements in AUROC over both reconstruction-based as well as adversarial state-of-the-art approaches to semi-supervised learning on low-resolution image benchmarks. The relative improvements in AUROC amounted to 13.0\%, 12.6\% and 8.0\% over Ladder Networks and 9.4\%, 10.4\% and 5.6\% over Feature Matching GANs at 25, 50 and 100 labels, respectively. We note that even Na\"ive Multitask Networks, that did not make use of any of the adaptions introduced by \themethod s, with the exception of two cases outperformed both Ladder Networks and Feature Matching GANs - suggesting that distant supervision in general is an efficacious approach to semi-supervised learning in this domain.

Interestingly, most of the evaluated semi-supervised approaches, with the exception of Na\"ive Multitask Networks, were outperformed by their purely supervised counterparts at lower amounts of labels than one would expect - in many cases by a large margin. Indeed, both Feature Matching GANs as well as Ladder Networks were eclipsed by the supervised baseline at just 100 labels. This suggests that either: (i) Feature Matching GANs and Ladder Networks require a higher degree of hyperparameter optimisation than the other evaluated approaches or (ii) the strengths of these approaches in the domain of low-resolution images do not generalise to the same degree to the domain of multivariate high-resolution time series without adaptations. These are novel findings given that most other recent evaluations of state-of-the-art methods in semi-supervised learning have been confined solely to the low-resolution image domain. We believe that, in the future, more systematic replication studies, such as the one presented in this work, are necessary to evaluate the degree to which new methods generalise beyond benchmark datasets that often do not cover many practically important data modalities, such as time series data, and idiosyncrasies, such as missingness, heterogeneity, sparsity and noise.

In terms of sensitivity and specificity, our best models would have been able to reduce the number of false alarms brought to the attention of clinical staff with the same degree of urgency as true alarms with sensitivities of 22.97\% (Feature Matching GAN), 40.99\% (\themethod-12), 48.76\% (\themethod-50), 63.60\% (\themethod-100R), 66.43\% (Feature RF) and 76.68\% (Feature RF) using, respectively, 12, 25, 50, 100, 500 and 1244 labelled training samples at a specificity of 95\%. In relative terms, \themethod s were therefore - with just 100 labels - able to realise $\frac{63.60}{76.68}=82.94\%$ of the expected reduction in false alarms of the Feature RF that was trained on 1,244 labels. This finding confirms that a modest data collection effort would be sufficient to achieve a considerable improvement in false alarm rates in critical care.

\textbf{Number of Auxiliary Tasks.} In \themethod s with auxiliary tasks selected by feature importance, more auxiliary tasks achieved slightly better performances once sufficient amounts of labels were available. We reason that, because the head block was trained on labelled samples only, a greater number of labels was necessary to effectively orchestrate the extra information provided by a larger number of multitask blocks. However, we did not see the same behavior in \themethod s with auxiliary tasks selected at random. Here, the performances of \themethod s with 6 and 100 auxiliary tasks were comparable across all label levels.

\textbf{Importance of Adaptions.} We found that using \themethod{}s trained with auxiliary tasks distributed in depth (\themethod-100D) performed worse than our proposed architecture - demonstrating that parallel alignment of multitask blocks is the superior architectural design choice. Similarly, \themethod{}-100 variants without the two step training procedure (- two step train) consistently failed to reach the semi-supervised performance of their counterparts with the two step training procedure enabled (\themethod-100) for more than 12 labels. This shows that disentangling the training of the auxiliary and the main task played an integral role in the strong semi-supervised performance of \themethod s and further reinforces prior reports that adverse gradient interactions are a key challenge for multitask learning in neural networks \cite{teh2017distral,doersch2017multi}.

\textbf{Task Selection.} We found that random selection in most cases outperformed selection in order of feature importance when comparing the \themethod-6 and \themethod-6R variants. We believe this was the result of increased task diversity when selecting at random from the relevant auxiliary tasks, as similar features rank close to each other in terms of feature importance. The fact that this effect was less pronounced between the same models with more auxiliary tasks (\themethod-100R and \themethod-100) supports this theory, as a larger set of tasks will automatically have a higher diversity due to the limited amount of highly similar features, thus decreasing the importance of accounting for diversity in the selection methodology. We therefore conclude that task diversity is the dominant factor in selecting related auxiliary tasks for distant multitask supervision.
\section{Limitations}
False alarms in the ICU are not solely a technical problem \cite{cvach2012monitor,drew2014insights}. Organisational and processual aspects must also be considered to comprehensibly address this issue in clinical care \cite{drew2014insights}. One such aspect is the question of how to best manage those alarms that have been flagged as false by an alarm classification system. We reason that, due to the inherent possibility of suppressing a true alarm, a sensible approach would be to report those errors with a lower degree of urgency, i.e. with a less pronounced sound, rather than completely suppressing them \cite{cvach2012monitor}. 

Another limitation of this work is that we only considered the detection of alarms that are caused by either artefacts or technical errors. Alarms that are technically correct, but clinically require no intervention, are another important source of false alarms \cite{drew2014insights} that we did not analyse in this work. Identifying clinically false alarms is significantly harder than those caused by artefacts and technical errors, as clinical reasoning requires deep knowledge of a patient's high-level physiological state, as well as a significant amount of domain knowledge.

Lastly, while the presented distantly supervised approach to semi-supervised learning performs well on our dataset, its applicability to other datasets hinges on being able to determine multiple related auxiliary tasks. We only evaluated distantly supervised multitask learning on time series data, where large numbers of suitable auxiliary tasks are readily available through automated feature extraction and selection \cite{christ2016distributed}. We hypothesise that it might not be trivial to find large repositories of auxiliary tasks suitable for distant multitask supervision for all data types. A comparatively small number of potential auxiliary tasks have been reported in related works in computer vision and natural language processing \cite{blaschko2010simultaneous,xu2015learning,oquab2015object,deriu2017leveraging,doersch2017multi}. Finally, our experiments yield insights into the importance of auxiliary task selection in \themethod s, but further theoretical analyses are necessary to understand exactly what types of auxiliary task are useful to what degree in distantly supervised multitask learning. 

\section{Conclusion}
We present a novel approach to reducing false alarms in the ICU using data obtained from a dynamic set of multiple heterogenous biosignal monitors. Unlabelled data is abundantly available, but obtaining trustworthy expert labels is laborious and expensive in this setting. We introduce a multitask network architecture that leverages distant supervision through multiple related auxiliary tasks in order to reduce the number of expensive labels required for training. We develop both a methodology for automatically selecting auxiliary tasks from multivariate time series as well as an optimised training procedure that counteracts adverse gradient interactions between tasks. Using a real-world critical care dataset, we demonstrate that our approach leads to significant improvements over several state-of-the-art baselines. In addition, we found that task diversity and adverse gradient interactions are key concerns in distantly supervised multitask learning. Going forward, we believe that our approach could be applicable to a wide variety of machine-learning tasks in healthcare for which obtaining labelled data is a major challenge.

\section*{Acknowledgements}
This work was partially funded by the Swiss National Science Foundation (SNSF) project No. 167195 within the National Research Program (NRP) $75$ ``Big Data'' and the Swiss Commission for Technology and Innovation (CTI) project No. 25531.

\small
\bibliography{example_paper}

\begin{thebibliography}{55}
\providecommand{\natexlab}[1]{#1}
\providecommand{\url}[1]{\texttt{#1}}
\expandafter\ifx\csname urlstyle\endcsname\relax
  \providecommand{\doi}[1]{doi: #1}\else
  \providecommand{\doi}{doi: \begingroup \urlstyle{rm}\Url}\fi

\bibitem[Aboukhalil et~al.(2008)Aboukhalil, Nielsen, Saeed, Mark, and
  Clifford]{aboukhalil2008reducing}
Aboukhalil, A., Nielsen, L., Saeed, M., Mark, R.~G., and Clifford, G.~D.
\newblock Reducing false alarm rates for critical arrhythmias using the
  arterial blood pressure waveform.
\newblock \emph{Journal of Biomedical Informatics}, 41\penalty0 (3):\penalty0
  442--451, 2008.

\bibitem[Ando \& Zhang(2005)Ando and Zhang]{ando2005framework}
Ando, R.~K. and Zhang, T.
\newblock A framework for learning predictive structures from multiple tasks
  and unlabeled data.
\newblock \emph{Journal of Machine Learning Research}, 6\penalty0
  (Nov):\penalty0 1817--1853, 2005.

\bibitem[Aytar et~al.(2018)Aytar, Pfaff, Budden, Paine, Wang, and
  de~Freitas]{aytar2018playing}
Aytar, Y., Pfaff, T., Budden, D., Paine, T.~L., Wang, Z., and de~Freitas, N.
\newblock Playing hard exploration games by watching youtube.
\newblock \emph{arXiv preprint arXiv:1805.11592}, 2018.

\bibitem[Baxter(2000)]{baxter2000model}
Baxter, J.
\newblock A model of inductive bias learning.
\newblock \emph{{Journal of Artificial Intelligence Research}}, 12\penalty0
  (149-198):\penalty0 3, 2000.

\bibitem[Ben-David \& Schuller(2003)Ben-David and Schuller]{ben2003exploiting}
Ben-David, S. and Schuller, R.
\newblock Exploiting task relatedness for multiple task learning.
\newblock \emph{{Lecture Notes in Computer Science}}, pp.\  567--580, 2003.

\bibitem[Blaschko et~al.(2010)Blaschko, Vedaldi, and
  Zisserman]{blaschko2010simultaneous}
Blaschko, M., Vedaldi, A., and Zisserman, A.
\newblock Simultaneous object detection and ranking with weak supervision.
\newblock In \emph{Advances in neural information processing systems}, pp.\
  235--243, 2010.

\bibitem[Che et~al.(2016)Che, Purushotham, Cho, Sontag, and
  Liu]{che2016recurrent}
Che, Z., Purushotham, S., Cho, K., Sontag, D., and Liu, Y.
\newblock Recurrent neural networks for multivariate time series with missing
  values.
\newblock \emph{arXiv preprint arXiv:1606.01865}, 2016.

\bibitem[Cheng et~al.(2017)Cheng, Darnell, Chivers, Draugelis, Li, and
  Engelhardt]{cheng2017sparse}
Cheng, L.-F., Darnell, G., Chivers, C., Draugelis, M.~E., Li, K., and
  Engelhardt, B.~E.
\newblock {Sparse multi-output Gaussian processes for medical time series
  prediction}.
\newblock \emph{arXiv preprint arXiv:1703.09112}, 2017.

\bibitem[Choi et~al.(2016)Choi, Bahadori, Schuetz, Stewart, and
  Sun]{choi2016doctor}
Choi, E., Bahadori, M.~T., Schuetz, A., Stewart, W.~F., and Sun, J.
\newblock {Doctor AI: Predicting clinical events via recurrent neural
  networks}.
\newblock In \emph{Machine Learning for Healthcare Conference}, pp.\  301--318,
  2016.

\bibitem[Christ et~al.(2016)Christ, Kempa-Liehr, and
  Feindt]{christ2016distributed}
Christ, M., Kempa-Liehr, A.~W., and Feindt, M.
\newblock Distributed and parallel time series feature extraction for
  industrial big data applications.
\newblock \emph{arXiv preprint arXiv:1610.07717}, 2016.

\bibitem[Clifford et~al.(2017)Clifford, Liu, Moody, Lehman, Silva, Li, Johnson,
  and Mark]{clifford2017af}
Clifford, G., Liu, C., Moody, B., Lehman, L., Silva, I., Li, Q., Johnson, A.,
  and Mark, R.~G.
\newblock {AF classification from a short single lead ECG recording: The
  Physionet Computing in Cardiology Challenge 2017}.
\newblock \emph{Computing in Cardiology}, 44, 2017.

\bibitem[Clifford et~al.(2015)Clifford, Silva, Moody, Li, Kella, Shahin,
  Kooistra, Perry, and Mark]{clifford2015physionet}
Clifford, G.~D., Silva, I., Moody, B., Li, Q., Kella, D., Shahin, A., Kooistra,
  T., Perry, D., and Mark, R.~G.
\newblock {The PhysioNet/Computing in Cardiology Challenge 2015: Reducing false
  arrhythmia alarms in the ICU}.
\newblock In \emph{Computing in Cardiology}, pp.\  273--276. IEEE, 2015.

\bibitem[Cvach(2012)]{cvach2012monitor}
Cvach, M.
\newblock {Monitor alarm fatigue: An integrative review}.
\newblock \emph{Biomedical Instrumentation \& Technology}, 46\penalty0
  (4):\penalty0 268--277, 2012.

\bibitem[Dai et~al.(2017)Dai, Yang, Yang, Cohen, and
  Salakhutdinov]{dai2017good}
Dai, Z., Yang, Z., Yang, F., Cohen, W.~W., and Salakhutdinov, R.
\newblock {Good Semi-supervised Learning that Requires a Bad GAN}.
\newblock \emph{{Advances in Neural Information Processing Systems}}, 2017.

\bibitem[Deriu et~al.(2017)Deriu, Lucchi, De~Luca, Severyn, M{\"u}ller,
  Cieliebak, Hofmann, and Jaggi]{deriu2017leveraging}
Deriu, J., Lucchi, A., De~Luca, V., Severyn, A., M{\"u}ller, S., Cieliebak, M.,
  Hofmann, T., and Jaggi, M.
\newblock Leveraging large amounts of weakly supervised data for multi-language
  sentiment classification.
\newblock In \emph{Proceedings of the 26th International Conference on World
  Wide Web}, pp.\  1045--1052. International World Wide Web Conferences
  Steering Committee, 2017.

\bibitem[Doersch \& Zisserman(2017)Doersch and Zisserman]{doersch2017multi}
Doersch, C. and Zisserman, A.
\newblock Multi-task self-supervised visual learning.
\newblock In \emph{Proceedings of the IEEE Conference on Computer Vision and
  Pattern Recognition}, pp.\  2051--2060, 2017.

\bibitem[Drew et~al.(2014)Drew, Harris, Z{\`e}gre-Hemsey, Mammone, Schindler,
  Salas-Boni, Bai, Tinoco, Ding, and Hu]{drew2014insights}
Drew, B.~J., Harris, P., Z{\`e}gre-Hemsey, J.~K., Mammone, T., Schindler, D.,
  Salas-Boni, R., Bai, Y., Tinoco, A., Ding, Q., and Hu, X.
\newblock {Insights into the problem of alarm fatigue with physiologic monitor
  devices: A comprehensive observational study of consecutive intensive care
  unit patients}.
\newblock \emph{PloS one}, 9\penalty0 (10):\penalty0 e110274, 2014.

\bibitem[Eerik{\"a}inen et~al.(2015)Eerik{\"a}inen, Vanschoren, Rooijakkers,
  Vullings, and Aarts]{eerikainen2015decreasing}
Eerik{\"a}inen, L.~M., Vanschoren, J., Rooijakkers, M.~J., Vullings, R., and
  Aarts, R.~M.
\newblock Decreasing the false alarm rate of arrhythmias in intensive care
  using a machine learning approach.
\newblock In \emph{Computing in Cardiology}, 2015.

\bibitem[Fallet et~al.(2015)Fallet, Yazdani, and Vesin]{fallet2015multimodal}
Fallet, S., Yazdani, S., and Vesin, J.-M.
\newblock A multimodal approach to reduce false arrhythmia alarms in the
  intensive care unit.
\newblock In \emph{Computing in Cardiology}, 2015.

\bibitem[Fernando et~al.(2017)Fernando, Bilen, Gavves, and
  Gould]{fernando2017self}
Fernando, B., Bilen, H., Gavves, E., and Gould, S.
\newblock Self-supervised video representation learning with odd-one-out
  networks.
\newblock In \emph{2017 IEEE Conference on Computer Vision and Pattern
  Recognition (CVPR)}, pp.\  5729--5738. IEEE, 2017.

\bibitem[Ghassemi et~al.(2014)Ghassemi, Naumann, Doshi-Velez, Brimmer, Joshi,
  Rumshisky, and Szolovits]{ghassemi2014unfolding}
Ghassemi, M., Naumann, T., Doshi-Velez, F., Brimmer, N., Joshi, R., Rumshisky,
  A., and Szolovits, P.
\newblock {Unfolding physiological state: Mortality modelling in intensive care
  units}.
\newblock In \emph{Proceedings of the 20th ACM SIGKDD International Conference
  on Knowledge Discovery and Data Mining}, pp.\  75--84. ACM, 2014.

\bibitem[Ghassemi et~al.(2015)Ghassemi, Pimentel, Naumann, Brennan, Clifton,
  Szolovits, and Feng]{ghassemi2015multivariate}
Ghassemi, M., Pimentel, M.~A., Naumann, T., Brennan, T., Clifton, D.~A.,
  Szolovits, P., and Feng, M.
\newblock {A multivariate timeseries modeling approach to severity of illness
  assessment and forecasting in ICU with sparse, heterogeneous clinical data}.
\newblock In \emph{Proceedings of the Twenty-Ninth AAAI Conference on
  Artificial Intelligence}, pp.\  446--453, 2015.

\bibitem[Goodfellow et~al.(2014)Goodfellow, Pouget-Abadie, Mirza, Xu,
  Warde-Farley, Ozair, Courville, and Bengio]{goodfellow2014generative}
Goodfellow, I., Pouget-Abadie, J., Mirza, M., Xu, B., Warde-Farley, D., Ozair,
  S., Courville, A., and Bengio, Y.
\newblock Generative adversarial nets.
\newblock In \emph{Advances in Neural Information Processing Systems}, pp.\
  2672--2680, 2014.

\bibitem[He et~al.(2016)He, Zhang, Ren, and Sun]{he2016deep}
He, K., Zhang, X., Ren, S., and Sun, J.
\newblock Deep residual learning for image recognition.
\newblock In \emph{Proceedings of the IEEE conference on computer vision and
  pattern recognition}, pp.\  770--778, 2016.

\bibitem[Jaderberg et~al.(2016)Jaderberg, Mnih, Czarnecki, Schaul, Leibo,
  Silver, and Kavukcuoglu]{jaderberg2016reinforcement}
Jaderberg, M., Mnih, V., Czarnecki, W.~M., Schaul, T., Leibo, J.~Z., Silver,
  D., and Kavukcuoglu, K.
\newblock Reinforcement learning with unsupervised auxiliary tasks.
\newblock \emph{arXiv preprint arXiv:1611.05397}, 2016.

\bibitem[Jordan \& Jacobs(1994)Jordan and Jacobs]{jordan1994hierarchical}
Jordan, M.~I. and Jacobs, R.~A.
\newblock {Hierarchical mixtures of experts and the EM algorithm}.
\newblock \emph{Neural computation}, 6\penalty0 (2):\penalty0 181--214, 1994.

\bibitem[Kendall(1945)]{kendall1945treatment}
Kendall, M.~G.
\newblock The treatment of ties in ranking problems.
\newblock \emph{Biometrika}, pp.\  239--251, 1945.

\bibitem[Kingma \& Welling(2014)Kingma and Welling]{kingma2014auto}
Kingma, D.~P. and Welling, M.
\newblock Auto-encoding variational bayes.
\newblock \emph{International Conference on Learning Representations}, 2014.

\bibitem[Kingma et~al.(2014)Kingma, Mohamed, Rezende, and
  Welling]{kingma2014semi}
Kingma, D.~P., Mohamed, S., Rezende, D.~J., and Welling, M.
\newblock Semi-supervised learning with deep generative models.
\newblock In \emph{Advances in Neural Information Processing Systems}, pp.\
  3581--3589, 2014.

\bibitem[Krasteva et~al.(2016)Krasteva, Jekova, Leber, Schmid, and
  Ab{\"a}cherli]{krasteva2016real}
Krasteva, V., Jekova, I., Leber, R., Schmid, R., and Ab{\"a}cherli, R.
\newblock Real-time arrhythmia detection with supplementary ecg quality and
  pulse wave monitoring for the reduction of false alarms in icus.
\newblock \emph{Physiological measurement}, 37\penalty0 (8):\penalty0 1273,
  2016.

\bibitem[Laine \& Aila(2017)Laine and Aila]{laine2016temporal}
Laine, S. and Aila, T.
\newblock Temporal ensembling for semi-supervised learning.
\newblock \emph{International Conference on Learning Representations}, 2017.

\bibitem[Lasko et~al.(2013)Lasko, Denny, and Levy]{lasko2013computational}
Lasko, T.~A., Denny, J.~C., and Levy, M.~A.
\newblock Computational phenotype discovery using unsupervised feature learning
  over noisy, sparse, and irregular clinical data.
\newblock \emph{PloS one}, 8\penalty0 (6):\penalty0 e66341, 2013.

\bibitem[Li et~al.(2017)Li, Xu, Zhu, and Zhang]{li2017triple}
Li, C., Xu, K., Zhu, J., and Zhang, B.
\newblock Triple generative adversarial nets.
\newblock \emph{{Advances in Neural Information Processing Systems}}, 2017.

\bibitem[Libbrecht \& Noble(2015)Libbrecht and Noble]{libbrecht2015machine}
Libbrecht, M.~W. and Noble, W.~S.
\newblock Machine learning applications in genetics and genomics.
\newblock \emph{Nature Reviews Genetics}, 16\penalty0 (6):\penalty0 321, 2015.

\bibitem[Lipton et~al.(2016{\natexlab{a}})Lipton, Kale, Elkan, and
  Wetzell]{lipton2015learning}
Lipton, Z.~C., Kale, D.~C., Elkan, C., and Wetzell, R.
\newblock {Learning to diagnose with LSTM recurrent neural networks}.
\newblock \emph{International Conference on Learning Representations},
  2016{\natexlab{a}}.

\bibitem[Lipton et~al.(2016{\natexlab{b}})Lipton, Kale, and
  Wetzel]{lipton2016directly}
Lipton, Z.~C., Kale, D.~C., and Wetzel, R.
\newblock {Directly modeling missing data in sequences with RNNs: Improved
  classification of clinical time series}.
\newblock In \emph{Machine Learning for Healthcare Conference}, pp.\  253--270,
  2016{\natexlab{b}}.

\bibitem[Oquab et~al.(2015)Oquab, Bottou, Laptev, and Sivic]{oquab2015object}
Oquab, M., Bottou, L., Laptev, I., and Sivic, J.
\newblock {Is object localization for free? Weakly-supervised learning with
  convolutional neural networks}.
\newblock In \emph{Proceedings of the IEEE Conference on Computer Vision and
  Pattern Recognition}, pp.\  685--694, 2015.

\bibitem[Papandreou et~al.(2015)Papandreou, Chen, Murphy, and
  Yuille]{papandreou2015weakly}
Papandreou, G., Chen, L.-C., Murphy, K.~P., and Yuille, A.~L.
\newblock Weakly-and semi-supervised learning of a deep convolutional network
  for semantic image segmentation.
\newblock In \emph{Proceedings of the IEEE international conference on computer
  vision}, pp.\  1742--1750, 2015.

\bibitem[Plesinger et~al.(2016)Plesinger, Klimes, Halamek, and
  Jurak]{plesinger2016taming}
Plesinger, F., Klimes, P., Halamek, J., and Jurak, P.
\newblock Taming of the monitors: reducing false alarms in intensive care
  units.
\newblock \emph{Physiological measurement}, 37\penalty0 (8):\penalty0 1313,
  2016.

\bibitem[Prasad et~al.(2017)Prasad, Cheng, Chivers, Draugelis, and
  Engelhardt]{prasad2017reinforcement}
Prasad, N., Cheng, L.-F., Chivers, C., Draugelis, M., and Engelhardt, B.~E.
\newblock A reinforcement learning approach to weaning of mechanical
  ventilation in intensive care units.
\newblock \emph{arXiv preprint arXiv:1704.06300}, 2017.

\bibitem[Ramsundar et~al.(2015)Ramsundar, Kearnes, Riley, Webster, Konerding,
  and Pande]{ramsundar2015massively}
Ramsundar, B., Kearnes, S., Riley, P., Webster, D., Konerding, D., and Pande,
  V.
\newblock Massively multitask networks for drug discovery.
\newblock \emph{arXiv preprint arXiv:1502.02072}, 2015.

\bibitem[Rasmus et~al.(2015)Rasmus, Berglund, Honkala, Valpola, and
  Raiko]{rasmus2015semi}
Rasmus, A., Berglund, M., Honkala, M., Valpola, H., and Raiko, T.
\newblock Semi-supervised learning with ladder networks.
\newblock In \emph{Advances in Neural Information Processing Systems}, pp.\
  3546--3554, 2015.

\bibitem[Saeed et~al.(2011)Saeed, Villarroel, Reisner, Clifford, Lehman, Moody,
  Heldt, Kyaw, Moody, and Mark]{saeed2011multiparameter}
Saeed, M., Villarroel, M., Reisner, A.~T., Clifford, G.~D., Lehman, L.-W.,
  Moody, G., Heldt, T., Kyaw, T.~H., Moody, B., and Mark, R.~G.
\newblock {Multiparameter Intelligent Monitoring in Intensive Care II
  (MIMIC-II): A public-access intensive care unit database}.
\newblock \emph{Critical Care Medicine}, 39\penalty0 (5):\penalty0 952, 2011.

\bibitem[Salimans et~al.(2016)Salimans, Goodfellow, Zaremba, Cheung, Radford,
  and Chen]{salimans2016improved}
Salimans, T., Goodfellow, I., Zaremba, W., Cheung, V., Radford, A., and Chen,
  X.
\newblock {Improved techniques for training GANs}.
\newblock In \emph{Advances in Neural Information Processing Systems}, pp.\
  2234--2242, 2016.

\bibitem[Saria et~al.(2010)Saria, Rajani, Gould, Koller, and
  Penn]{saria2010integration}
Saria, S., Rajani, A.~K., Gould, J., Koller, D., and Penn, A.~A.
\newblock Integration of early physiological responses predicts later illness
  severity in preterm infants.
\newblock \emph{Science translational medicine}, 2\penalty0 (48):\penalty0
  48ra65--48ra65, 2010.

\bibitem[Schwab et~al.(2017)Schwab, Scebba, Zhang, Delai, and
  Karlen]{schwab2017beat}
Schwab, P., Scebba, G.~C., Zhang, J., Delai, M., and Karlen, W.
\newblock {Beat by Beat: Classifying Cardiac Arrhythmias with Recurrent Neural
  Networks}.
\newblock In \emph{{Computing in Cardiology}}, 2017.

\bibitem[Schwab et~al.(2018)Schwab, Miladinovic, and Karlen]{schwab2018granger}
Schwab, P., Miladinovic, D., and Karlen, W.
\newblock Granger-causal attentive mixtures of experts: Learning important
  features with neural networks.
\newblock \emph{arXiv preprint arXiv:1802.02195}, 2018.

\bibitem[Shazeer et~al.(2017)Shazeer, Mirhoseini, Maziarz, Davis, Le, Hinton,
  and Dean]{shazeer2017outrageously}
Shazeer, N., Mirhoseini, A., Maziarz, K., Davis, A., Le, Q., Hinton, G., and
  Dean, J.
\newblock {Outrageously large neural networks: The sparsely-gated
  mixture-of-experts layer}.
\newblock \emph{arXiv preprint arXiv:1701.06538}, 2017.

\bibitem[Springenberg(2015)]{springenberg2015unsupervised}
Springenberg, J.~T.
\newblock Unsupervised and semi-supervised learning with categorical generative
  adversarial networks.
\newblock \emph{arXiv preprint arXiv:1511.06390}, 2015.

\bibitem[Srivastava et~al.(2015)Srivastava, Greff, and
  Schmidhuber]{srivastava2015training}
Srivastava, R.~K., Greff, K., and Schmidhuber, J.
\newblock Training very deep networks.
\newblock In \emph{{Advances in Neural Information Processing Systems}}, pp.\
  2377--2385, 2015.

\bibitem[Teh et~al.(2017)Teh, Bapst, Pascanu, Heess, Quan, Kirkpatrick,
  Czarnecki, and Hadsell]{teh2017distral}
Teh, Y., Bapst, V., Pascanu, R., Heess, N., Quan, J., Kirkpatrick, J.,
  Czarnecki, W.~M., and Hadsell, R.
\newblock Distral: Robust multitask reinforcement learning.
\newblock In \emph{Advances in Neural Information Processing Systems}, pp.\
  4497--4507, 2017.

\bibitem[Vincent et~al.(2008)Vincent, Larochelle, Bengio, and
  Manzagol]{vincent2008extracting}
Vincent, P., Larochelle, H., Bengio, Y., and Manzagol, P.-A.
\newblock Extracting and composing robust features with denoising autoencoders.
\newblock In \emph{{Proceedings of the 25th International Conference on Machine
  Learning}}, pp.\  1096--1103. ACM, 2008.

\bibitem[Wiens et~al.(2016)Wiens, Guttag, and Horvitz]{wiens2016patient}
Wiens, J., Guttag, J., and Horvitz, E.
\newblock {Patient risk stratification with time-varying parameters: A
  multitask learning approach}.
\newblock \emph{The Journal of Machine Learning Research}, 17\penalty0
  (1):\penalty0 2797--2819, 2016.

\bibitem[Xu et~al.(2015)Xu, Schwing, and Urtasun]{xu2015learning}
Xu, J., Schwing, A.~G., and Urtasun, R.
\newblock Learning to segment under various forms of weak supervision.
\newblock In \emph{Computer Vision and Pattern Recognition (CVPR), 2015 IEEE
  Conference on}, pp.\  3781--3790. IEEE, 2015.

\bibitem[Zeng et~al.(2015)Zeng, Liu, Chen, and Zhao]{zeng2015distant}
Zeng, D., Liu, K., Chen, Y., and Zhao, J.
\newblock Distant supervision for relation extraction via piecewise
  convolutional neural networks.
\newblock In \emph{Proceedings of the 2015 Conference on Empirical Methods in
  Natural Language Processing}, pp.\  1753--1762, 2015.

\end{thebibliography}
\bibliographystyle{icml2018}



\section*{Supplementary material}

# Supplementary Material for:
## "Not to Cry Wolf:
## Distantly Supervised Multitask Learning in Critical Care"

**Patrick Schwab** [1]  **Emanuela Keller** [2]  **Carl Muroi** [2]  **David J. Mack** [2]  **Christian Strässle** [2]  **Walter Karlen** [1]

## 1. Source Code

The source code for this work is available online at https://github.com/d909b/DSMT-Nets.

## 2. Instructions for Annotators

We instructed our annotators to label a given alarm context window as caused by an artefact if:

1. The signal that caused the alarm is not being recorded, as verified by visibility on the monitor.

2. The alarm-generating signal curve has an atypical shape.

3. Numerical values derived from the alarm-generating signal are not physiologically plausible.

Figures S1 and S2 depict qualitative examples of context windows that have been labelled as caused by an artefact.

---

[1]Institute of Robotics and Intelligent Systems, ETH Zurich, Switzerland [2]Neurocritical Care Unit, Department of Neuro-surgery, University Hospital Zurich, Switzerland. Correspondence to: Patrick Schwab <patrick.schwab@hest.ethz.ch>.

*Table S1.* Comparison of the standard deviation of AUROC values across the 35 distinct models (vertical) that we trained using different sets of hyperparameters and varying amounts of labels (horizontal). We report the AUROC of the best encountered model as calculated on the test set of 533 alarms. The worst result in each column is highlighted in bold. A higher variation in AUROC across hyperparameter choices and training runs may indicate higher sensitivity to hyperparameters in the evaluated range and/or lacking robustness of training in the presented setting. Most notably, we find that disentangling training of the auxiliary and the main task in DSMT-Nets improves training stability in most cases.

| AUROC with # of Labels | 12 | 25 | 50 | 100 | 500 | 1244 |
|---|---|---|---|---|---|---|
| Feature RF | - | - | - | - | - | - |
| Supervised baseline | 0.055 | 0.046 | 0.045 | 0.026 | 0.008 | 0.007 |
| Naïve Multitask Network | 0.061 | 0.057 | 0.048 | 0.054 | 0.049 | 0.041 |
| Ladder Network | 0.067 | 0.074 | 0.069 | 0.076 | **0.066** | **0.076** |
| Feature Matching GAN | 0.050 | 0.051 | 0.051 | 0.037 | 0.020 | 0.027 |
| | | | | | | |
| DSMT-Net-6 | 0.059 | 0.058 | 0.056 | 0.070 | 0.040 | 0.041 |
| DSMT-Net-12 | 0.058 | 0.064 | 0.072 | 0.074 | 0.040 | 0.037 |
| DSMT-Net-25 | 0.066 | 0.062 | 0.068 | 0.076 | 0.038 | 0.043 |
| DSMT-Net-50 | 0.059 | 0.071 | 0.066 | 0.060 | 0.042 | 0.044 |
| DSMT-Net-100 | 0.060 | 0.060 | **0.075** | 0.048 | 0.032 | 0.039 |
| - two step train | **0.070** | **0.076** | 0.065 | **0.078** | 0.058 | 0.061 |
| | | | | | | |
| DSMT-Net-6R | 0.058 | 0.051 | 0.061 | 0.062 | 0.039 | 0.035 |
| DSMT-Net-100R | 0.070 | 0.062 | 0.054 | 0.047 | 0.034 | 0.048 |
| DSMT-Net-100D | 0.019 | 0.023 | 0.038 | 0.021 | 0.030 | 0.038 |

*Table S2.* Comparison of the minimum AUROC value across the 35 distinct models (vertical) that we trained using different sets of hyperparameters and varying amounts of labels (horizontal). We report the AUROC of the best encountered model as calculated on the test set of 533 alarms. The best results in each column are highlighted in bold. The difference between the maximum and minimum value indicates the range of values covered over the 35 hyperparameter settings.

| AUROC with # of Labels | 12 | 25 | 50 | 100 | 500 | 1244 |
|---|---|---|---|---|---|---|
| Feature RF | - | - | - | - | - | - |
| Supervised baseline | 0.501 | 0.547 | 0.568 | **0.763** | **0.907** | **0.911** |
| Naïve Multitask Network | 0.516 | 0.577 | 0.613 | 0.648 | 0.693 | 0.732 |
| Ladder Network | 0.506 | 0.516 | 0.538 | 0.512 | 0.594 | 0.560 |
| Feature Matching GAN | **0.629** | **0.628** | 0.646 | 0.719 | 0.817 | 0.757 |
| | | | | | | |
| DSMT-Net-6 | 0.514 | 0.557 | 0.588 | 0.604 | 0.760 | 0.752 |
| DSMT-Net-12 | 0.507 | 0.540 | 0.579 | 0.630 | 0.753 | 0.791 |
| DSMT-Net-25 | 0.501 | 0.603 | 0.535 | 0.570 | 0.774 | 0.779 |
| DSMT-Net-50 | 0.506 | 0.557 | 0.649 | 0.682 | 0.768 | 0.770 |
| DSMT-Net-100 | 0.507 | 0.552 | 0.600 | 0.691 | 0.797 | 0.774 |
| - two step train | 0.502 | 0.500 | 0.539 | 0.525 | 0.645 | 0.685 |
| | | | | | | |
| DSMT-Net-6R | 0.515 | 0.624 | 0.630 | 0.635 | 0.760 | 0.805 |
| DSMT-Net-100R | 0.506 | 0.601 | **0.660** | 0.686 | 0.771 | 0.771 |
| DSMT-Net-100D | 0.500 | 0.500 | 0.500 | 0.500 | 0.500 | 0.500 |

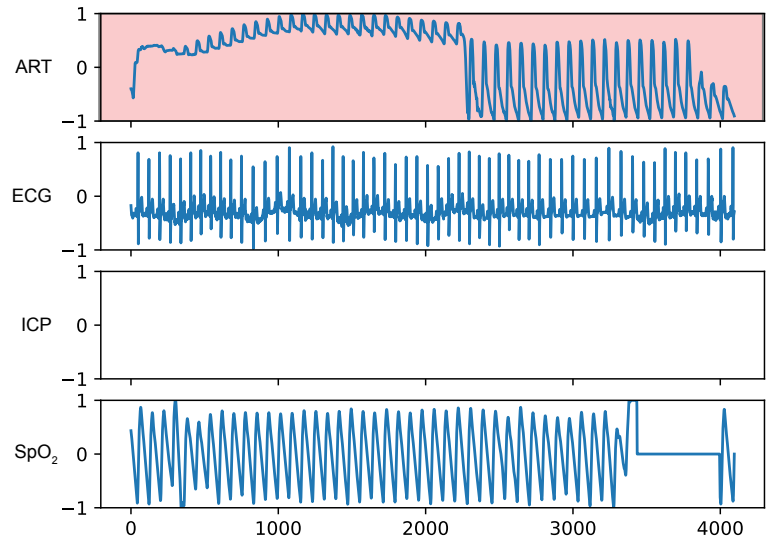

*Figure S1.* A qualitative example of an alarm caused by an artefact, as encountered in the ICU dataset. Depicted are the amplitudes (y-axis, standardised) over time (x-axis, in hundredths of a second) of the arterial blood pressure (ART), electrocardiography (ECG), intracranial pressure (ICP) and pulse oximetry ($SpO_2$) signals immediately before the alarm was triggered. An empty box indicates a missing signal. In this case, the alarm was triggered by the arterial blood pressure monitor (red). Note that there also appears to be an artefact in the pulse oximetry signal that might have triggered another independent alarm concurrently.

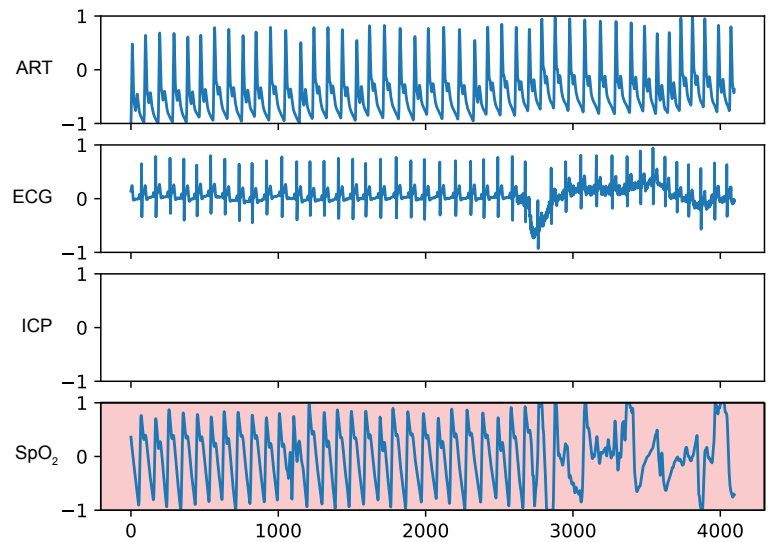

*Figure S2.* A qualitative example of an alarm caused by an artefact, as encountered in the ICU dataset. Depicted are the amplitudes (y-axis, standardised) over time (x-axis, in hundredths of a second) of the arterial blood pressure (ART), electrocardiography (ECG), intracranial pressure (ICP) and pulse oximetry ($SpO_2$) signals immediately before the alarm was triggered. An empty box indicates a missing signal. In this case, the alarm was triggered by the pulse oximetry monitor (red).

*Table S3.* The exact hyperparameter values used for each model for each of the 35 distinct training runs. We chose the values using a uniformly random selection within the ranges specified in the main paper. The number of hidden units per layer and the number of hidden layers were rounded to the nearest integer in our experiments.

| Run | Dropout | Number of hidden units / layer | Number of hidden layers |
|---|---|---|---|
| 1 | 0.5256 | 18.3015 | 1.4562 |
| 2 | 0.2926 | 26.3799 | 1.6650 |
| 3 | 0.3888 | 29.7946 | 1.4185 |
| 4 | 0.4633 | 29.1221 | 1.7195 |
| 5 | 0.3619 | 27.6884 | 1.7030 |
| 6 | 0.5049 | 26.5369 | 2.7647 |
| 7 | 0.7134 | 26.2866 | 1.4111 |
| 8 | 0.4486 | 23.7360 | 2.4363 |
| 9 | 0.2939 | 24.0741 | 1.1734 |
| 10 | 0.5652 | 21.2195 | 1.2685 |
| 11 | 0.3688 | 18.8924 | 2.5907 |
| 12 | 0.7542 | 20.2902 | 2.7300 |
| 13 | 0.2614 | 27.6143 | 1.5102 |
| 14 | 0.3820 | 24.7860 | 2.1281 |
| 15 | 0.3452 | 25.3250 | 2.9806 |
| 16 | 0.7308 | 30.3649 | 1.4315 |
| 17 | 0.6195 | 22.6811 | 1.7044 |
| 18 | 0.6170 | 21.3986 | 2.7229 |
| 19 | 0.7451 | 27.8114 | 2.2333 |
| 20 | 0.3469 | 22.9611 | 1.4900 |
| 21 | 0.5168 | 16.2036 | 2.9124 |
| 22 | 0.4098 | 20.5713 | 2.4480 |
| 23 | 0.3012 | 24.5169 | 1.3481 |
| 24 | 0.4475 | 17.3175 | 2.8138 |
| 25 | 0.2660 | 27.0517 | 1.2606 |
| 26 | 0.4830 | 21.8282 | 2.9766 |
| 27 | 0.7799 | 18.0746 | 2.1824 |
| 28 | 0.3712 | 24.3822 | 2.1989 |
| 29 | 0.5958 | 25.3871 | 2.8844 |
| 30 | 0.2649 | 30.3633 | 2.6249 |
| 31 | 0.6065 | 20.6158 | 1.9874 |
| 32 | 0.4623 | 16.1852 | 1.3220 |
| 33 | 0.2592 | 24.9682 | 1.8996 |
| 34 | 0.6531 | 26.4506 | 2.3409 |
| 35 | 0.7825 | 28.5137 | 2.9273 |



\end{document}